# Conditional Diffusion Modeling with Attention for Probabilistic Battery Capacity Prediction under Real-World Conditions

Chunlin Jiang[1], Hequn Li[1], Zhongwei Deng[2,*], Jie Shao[1], Zhansheng Ning[3,*]

[1] School of Computer Science and Engineering, University of Electronic Science and Technology of China, Chengdu, 611731, China

[2] School of Mechanical and Electrical Engineering, University of Electronic Science and Technology of China, Chengdu, 611731, China

[3] Faculty of Electrical Engineering, Mathematics and Computer Science, University of Twente, 7522 NB, Enschede, The Netherlands

[*]Corresponding Author: Prof. Deng, dengzw1127@uestc.edu.cn; Dr. Ning, z.ning@utwente.nl

**Abstract**

Accurate prediction of lithium-ion battery capacity and its associated uncertainty is essential for reliable battery management but remains challenging due to the stochastic nature of aging. This paper presents a new method, termed the Conditional Diffusion U-Net with Attention (CDUA), which integrates feature engineering and deep learning to address this challenge. The proposed approach employs a diffusion-based generative model for time-series forecasting and incorporates attention mechanisms to enhance predictive performance. Battery capacity is first derived from real-world vehicle operation data. The most relevant features are then identified using the Pearson correlation coefficient and the XGBoost algorithm. These features are used to train the CDUA model, which comprises two components: (1) a contextual U-Net with self-attention to capture complex temporal dependencies, and (2) a noise predictor network that learns to estimate the added noise, enabling the reconstruction of accurate capacity values from noisy observations. Experimental validation on the real-world vehicle data demonstrates that the proposed CDUA model achieves a relative mean absolute error of 0.94% and a relative root mean square error of 1.14%, with a narrow 95% confidence interval of 3.74% in relative width. These results confirm that CDUA provides both accurate capacity estimation and reliable uncertainty quantification. Comparative experiments further verify its robustness and superior performance over existing mainstream approaches.

## 1. Introduction

Electric vehicles (EVs) have been widely recognized as a viable solution to address the challenges of fossil fuel depletion and greenhouse gas emissions [1, 2]. Owing to their high energy density, environmental friendliness, and long service life, lithium-ion batteries have been adopted in numerous fields and deployed on a large scale, particularly in the EV industry [3, 4]. However, they pose significant safety risks and degrade over time, leading to capacity loss and an increased risk of failure [5, 6]. The failure of these batteries can result in performance degradation and even catastrophic consequences, such as fires or explosions, especially in EVs [7, 8]. In practical applications, battery aging leads to significant changes in key parameters, such as maximum available capacity and internal resistance [9, 10]. Therefore, achieving accurate battery capacity prediction is crucial for ensuring the safe, efficient, and reliable operation of EVs [11, 12].

Battery capacity prediction methods can be categorized into measurement-based analytical methods, model-driven methods, and data-driven methods [13]. Measurement-based analytical methods are further divided into direct measurement and indirect analysis. Direct measurement methods, including capacity tests [14], energy tests [15], and impedance tests [16, 17], provide precise data and highly reliable results. However, their reliance on specific experimental equipment and strict environmental conditions makes online, real-time monitoring difficult to implement. Indirect analytical methods include charge curve analysis [18, 19], incremental capacity analysis (ICA) [20], and differential voltage analysis (DVA) [21]. The advantage of these methods lies in their ability to intuitively characterize electrochemical aging mechanisms without complex models; yet, they are limited by the extremely high precision required for voltage and capacity measurements, and effective data acquisition under dynamic operating conditions is difficult. Model-based methods are classified into equivalent circuit models (ECMs) and electrochemical models. In the ECM framework, electronic components are used to simulate battery operation [22], while electrochemical models [23] use physical and chemical equations to describe the battery's electrochemical behavior. Although model-based methods are rooted in physical principles, their application is severely limited by the high difficulty of parameter identification and their generally poor generalization capability across different batteries [24]. In contrast, data-driven methods establish predictive models by analyzing large volumes of historical data generated during battery operation, significantly reducing reliance on electrochemical mechanisms [25, 26]. These methods offer good adaptability and real-time performance, enabling them to capture complex nonlinear degradation patterns from operational data. In recent years, with the widespread application of machine learning, data-driven approaches have become the mainstream for battery capacity prediction. By employing machine learning or deep learning algorithms, these methods automatically mine hidden patterns and features from massive battery aging datasets, enabling accurate capacity prediction without a deep understanding of microscopic aging mechanisms. Commonly used techniques include Support Vector Machine (SVM) [27], Random Forest (RF) [28], Convolutional Neural Network (CNN) [29], Recurrent Neural Network (RNN) [30, 31], and its variants like Long Short-Term Memory (LSTM) [32, 33] and Gated Recurrent Unit (GRU) [34], as well as Graph Neural Network (GNN) [35]. Among these, LSTM and GRU effectively mitigate the gradient problems of traditional RNNs through structural optimizations and gating mechanisms, enhancing their capability for modeling long-sequence data.

Although the aforementioned machine learning and deep learning models have achieved significant success in battery capacity prediction, they are fundamentally deterministic. These models typically provide a point prediction of a single value for the remaining battery capacity, making it difficult to quantify the uncertainty and confidence level of the prediction. This is a critical limitation, as battery aging is inherently a process fraught with stochastic noise and uncertainty. To overcome this limitation and provide a more comprehensive probabilistic distribution for capacity predictions, this study explores a generative approach based on diffusion models.

In recent years, generative models based on diffusion processes have garnered significant attention in the field of artificial intelligence. These models, which include approaches such as score matching [36] and Langevin dynamics [37], are valued for their ability to generate high-quality, high-dimensional data samples and ensure a stable training process. In particular, the framework of diffusion probabilistic models [38] has provided a more unified understanding of these generative models. The core idea lies in a parameterized Markov chain: a forward process that gradually transforms data into noise, and a reverse process that learns to reconstruct the data from that noise.

Although diffusion models originated in computer vision for generating high-dimensional image data, their application to predicting lithium-ion battery time series data presents certain challenges. Nevertheless, leveraging their superior generative capabilities and advantages in quantifying uncertainty, this study introduces the Conditional Diffusion U-Net with Attention (CDUA) model for lithium-ion battery capacity prediction. This approach can effectively characterize the inherent probabilistic nature of the stochastic process of battery aging, thereby providing a new and powerful solution for modeling lithium-ion battery capacity.

To address this challenge and fully leverage the advantages of diffusion models in probabilistic forecasting, this paper proposes a new method for lithium-ion battery capacity prediction based on CDUA. This study aims to transform the generative capabilities of diffusion models into precise probabilistic predictions of future battery capacity degradation, achieving both high-precision forecasting and robust uncertainty quantification through the integration of several advanced techniques. The main contributions of this paper include,

(1) A probabilistic forecasting framework based on diffusion models is proposed. Its core idea is to treat lithium-ion battery capacity prediction as a denoising process. By learning the patterns of capacity degradation from real-world vehicle data, the framework achieves probabilistic forecasting of future capacity. This approach not only provides accurate point prediction but also quantifies uncertainty for risk assessment, offering a new and reliable method for battery health management.

(2) A self-attention enhanced technique is implemented for small-sample scenarios. The core of the CDUA model is a U-Net architecture integrated with a self-attention mechanism. This enables the model to capture complex sequential dependencies and inter-feature correlations from limited real-world vehicle data (20 vehicles in this study), significantly enhancing its generalization capability in small-sample contexts.

(3) A general feature engineering framework is constructed. The framework combines statistical analysis and machine learning methods to efficiently screen for critical features from real-world vehicle charging data that are essential for battery capacity prediction. This hybrid selection strategy ensures high-quality input data, laying a solid foundation for the model's high-precision forecasting.

The remainder of this paper is organized as follows: Section II introduces the collection process for the real-world vehicle data, the calculation method for battery capacity labels, and the selection strategy for key features. Section III elaborates on the proposed CDUA model, detailing the design principles and implementation of its key modules. Section IV analyzes the performance of the proposed model in terms of point prediction and confidence interval prediction for battery capacity, compares it against other baseline models, and presents feature and model ablation studies to validate the effectiveness of each component. Finally, Section V concludes the paper with a summary of the findings and provides an outlook on future research directions.

## 2. Data preparation and feature engineering

This study utilizes the real-world vehicle charging datasets from our previous work [39]. The dataset comprises 20 vehicles (#1, #2, ..., #20) equipped with identical battery systems, and the operational period spans 120 weeks. The data was collected by charging devices, which received battery charging information via the Controller Area Network (CAN) during the charging process, with a recording interval of eight seconds. The collected data includes parameters such as timestamp, charging current, pack voltage, State of Charge (SOC), maximum/minimum cell voltage, and maximum/minimum temperature. Fig. 1(a-d) illustrates the key characteristics of current, voltage, SOC, and temperature during multiple charging segments, which are extracted and concatenated from the operational data of a single vehicle. The current profile in Fig. 1(a) exhibits a typical step-wise pattern, clearly indicating that the vehicle utilizes a multi-stage constant current charging strategy. Furthermore, the highly consistent voltage curves Fig. 1(b) and the well-controlled temperature curves Fig. 1(d) jointly demonstrate the battery pack's excellent cell balancing and effective thermal management.

### 2.1 Labeled capacity calculation and feature engineering

To enable the prediction of battery capacity, it is necessary to calculate the ground-truth values of the capacity. For each vehicle, the independent charging processes are first isolated, using a time interval greater than 10 seconds between data points as the criterion for separation. To ensure data quality, only segments with more than 100 data points and a stable SOC transition are retained for subsequent processing. For each valid charging segment, the Coulomb counting method is used to estimate the effective charge capacity:

$$C = \frac{-\int_{t_1}^{t_2} \Delta t I(t)}{SOC_{t_2} - SOC_{t_1}} \tag{1}$$

where $\Delta t$ is the fixed sampling interval, $I$ is the battery current (negative during charging), and $t_1$, $t_2$ are the start and end times of the charge, respectively.

However, obtaining an accurate SOC is challenging due to the lack of dedicated test data and key battery characteristics. To address this issue, this study aggregates the data on a weekly basis. The final labeled capacity for each week is determined by calculating the mean of the estimated capacities from all valid charging segments within that week. This averaging process effectively mitigates inaccuracies caused by individual SOC prediction errors. In this process, in addition to battery capacity, other data dimensions are also computed. This includes the mean, sum, and standard deviation of key operational parameters during charging (such as maximum/minimum temperature, maximum/minimum cell voltage, and pack voltage), resulting in a total of 27 features. Finally, all calculated features are aggregated on a weekly basis to facilitate the analysis of the capacity degradation trend, resulting in the battery capacity degradation curves for 20 vehicles as shown Fig. 1(e). As the total number of aggregated weeks differs for each vehicle, the data for all vehicles has been truncated to the shortest duration to allow for more intuitive observation and comparison.

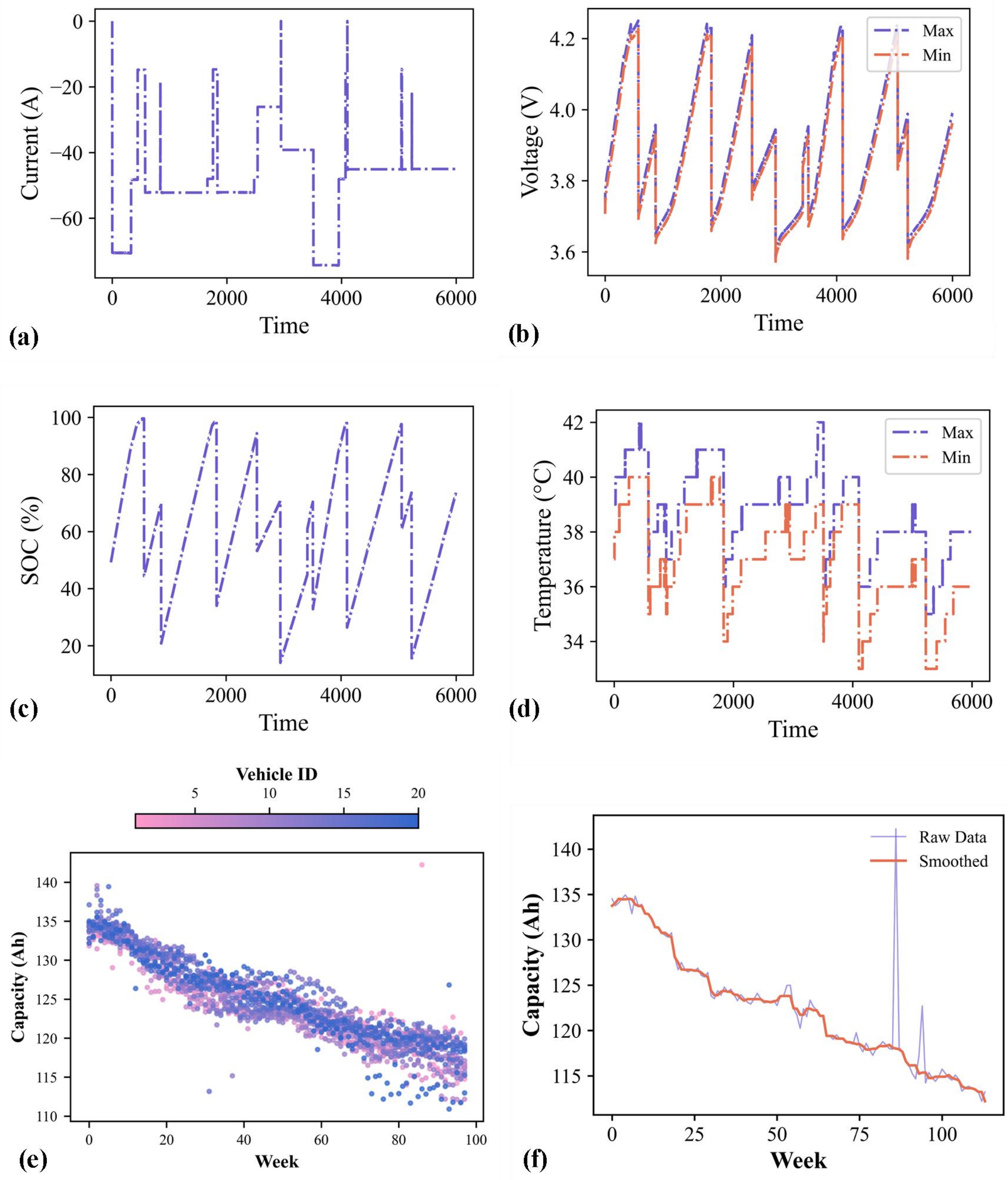

Fig. 1. Analysis of vehicle battery operating conditions and capacity degradation characteristics.(a) charging current curve; (b) charging voltage curve; (c) SOC variation curve during charging; (d) temperature variation during charging; (e) comparison of battery capacity degradation for 20 vehicles; (f) effect of median filtering on a noisy capacity trajectory.

### 2.2 Noise reduction using median filtering

During the data processing, outliers are observed in specific vehicle-week data segments. To smooth the data, this study employs a median filtering technique for noise reduction. This method utilizes a sliding window of an odd size and replaces the original data point with the local median within that window. This approach effectively eliminates outliers while preserving the overall characteristics of the time series data. Fig. 1(f) shows a comparison of the results before and after processing, where the filtered trajectory for vehicle #19 (in the 87th week) exhibits a significantly smoother profile compared to the original values.

### 2.3 Hybrid feature selection strategy

To enhance the correlation between the calculated features and the target capacity, a hybrid feature selection method is proposed, which combines Pearson correlation coefficients with XGBoost-based importance ranking. The Pearson method quantifies the static linear correlation between features and capacity, while XGBoost assesses the dynamic non-linear contribution of each feature to the degradation trend. Screening features using these complementary metrics yields a stronger overall feature-target correlation, thereby improving model accuracy. The Pearson correlation coefficient is calculated as follows:

$$\rho_{x,z} = \frac{\sum (x_i - \overline{x})(z_i - \overline{z})}{\sqrt{\sum (x_i - \overline{x})^2 \sum (z_i - \overline{z})^2}} \tag{2}$$

where $x_i$ is the value of the $i_{th}$ feature, $z$ is the battery pack capacity, and $\overline{x}$ , $\overline{z}$ are the mean values of the specific feature and capacity sequences, respectively.

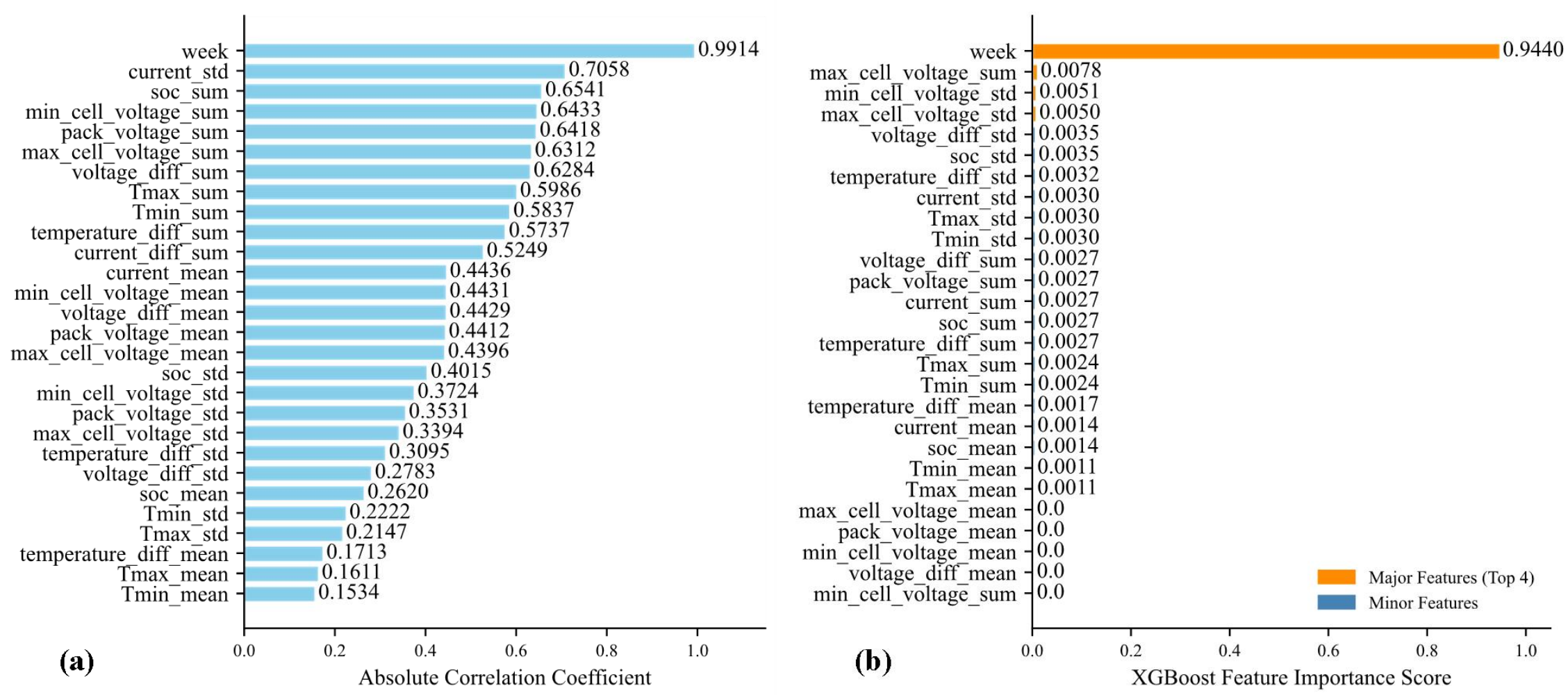

Fig. 2. Comparative analysis of feature importance: Pearson correlation vs. XGBoost. (a) Pearson correlation analysis of features with target value. (b) feature importance ranking by XGBoost model.

To visually demonstrate the relevance of each feature to the lithium-ion battery capacity, the selection results from both methods are presented in Fig. 2(a) and Fig. 2(b), where the absolute correlation value of each feature with the target capacity is visualized. Based on this analysis, features with a Pearson correlation score exceeding 0.6 (6 features in total) are selected to form set F1, and features with an XGBoost importance score greater than 0.01 (4 features in total) are selected for set F2. After removing the duplicate "Week" feature, the union of these two subsets results in a final set of 9 distinct features, denoted as F3. This final feature set includes: Week, Mean of Minimum Cell Voltage, Mean of Pack Voltage, Standard Deviation of Current, Sum of SOC, Sum of Pack Voltage, Sum of Minimum Cell Voltage, Sum of Maximum Cell Voltage, and Mean of Voltage Difference.

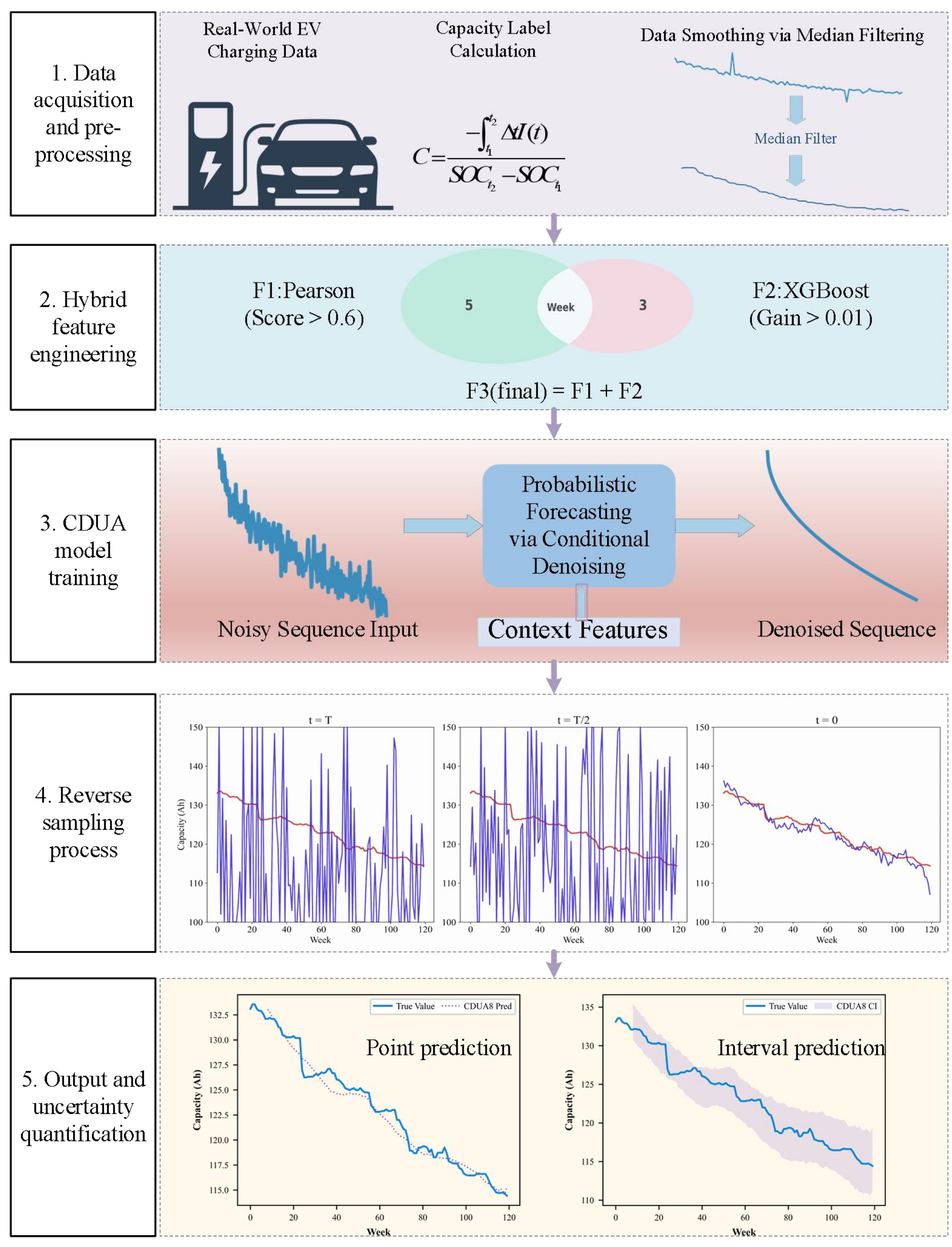

$$C = \frac{-\int_{t_1}^{t_2} \Delta I(t)}{SOC_{t_2} - SOC_{t_1}}$$

Fig. 3. Schematic representation of the proposed method.

## 3. Methodology

To address the challenges of randomness and uncertainty in predicting future capacity sequences of lithium-ion batteries, and to generate accurate prediction results, a conditional diffusion model based on U-Net and attention mechanism is proposed and named CDUA. The core idea of this model is to leverage the powerful generative capabilities of a diffusion model to generate a probability distribution for future capacity sequences. This is achieved by progressively denoising the data, guided by features from historical data, which enables the quantification of uncertainty.

This section details the composition and operational principles of the CDUA model, including the mathematical theory of the diffusion process, the overall architecture of the model, the design of key modules, as well as the training and inference strategies. To provide a clear and conceptual overview of our entire workflow, from raw data processing to the final probabilistic forecast, a schematic representation of the proposed methodology is presented in Fig. 3. The process begins with the acquisition of real-world EV charging data. Capacity labels are calculated and then smoothed via median filtering. Following this, a hybrid feature extraction strategy is employed, combining Pearson correlation and XGBoost importance scores to form the final feature set. These features serve as the conditional input for the core CDUA model. The model's central task is conceptualized as a conditional noise prediction problem, where it learns to transform a noisy sequence into a clean prediction by accurately estimating and removing noise, guided by context features. During inference, the model performs probabilistic forecasting through a reverse sampling process. This is visualized by generating multiple distinct future trajectories from random noise to form a predictive distribution. Finally, these trajectories are aggregated to produce the output, which includes a mean point forecast and the quantified uncertainty represented by a 95% confidence interval. The subsequent sections elaborate on these components in detail, beginning with the theoretical underpinnings of the conditional diffusion model.

### 3.1 Conditional diffusion model theory

The CDUA model is built upon the foundation of Denoising Diffusion Probabilistic Models (DDPM). DDPM comprises two core processes: a fixed forward diffusion process and a learnable reverse denoising process. The forward diffusion process is a fixed Markov chain that, over T discrete time steps, gradually adds Gaussian noise to a true future battery capacity sequence *y*0. This process continues until the sequence is ultimately transformed into a pure noise sequence following a standard Gaussian distribution. This process is defined as:

$$q(y_t \mid y_{t-1}) \coloneqq \mathcal{N}(\sqrt{1-\beta_t}\, y_{t-1}, \beta_t I) \tag{3}$$

where $\beta_t$ represents the noise variance at time step *t*. For this model, a linear noise scheduling strategy is adopted, where the value of $\beta_t$ increases linearly from $\beta_1$ = 0.0001 to $\beta_2$ = 0.02.

A key feature of the forward process is that by using a reparameterization trick, we can directly sample the noisy data $y_t$ at any arbitrary time step *t* from the initial data $y_0$ . This is made possible by defining $\bar{\alpha}_t = 1-\beta_t$ and $\alpha_t = \prod_{s=1}^{t} \bar{\alpha}_s$ .

The noisy sample can then be directly expressed as:

$$q(y_t \mid y_0) = \mathcal{N}(\sqrt{\alpha_t}\, y_0, (1-\alpha_t) I) \tag{4}$$

This formula forms the basis for the model's training, as it allows us to efficiently sample noisy data $y_t$ and its corresponding target noise $\epsilon$, at any time step $t$. This eliminates the need to run the sequential Markov chain for each training sample, significantly speeding up the training process.

The reverse denoising process is the core of the entire diffusion model. It is a Markov chain that starts from a pure noise sample $y_T$, and progressively predicts and removes the noise to ultimately recover the true data $y_0$. To visually illustrate this process, Fig. 4 shows several key timesteps, starting from a pure noise sample and progressively generating a clean battery capacity prediction curve. In the figure, the blue line in each subplot represents the noisy sample $y_t$ at that particular moment in the process. The thick red line, on the other hand, represents the model's estimate of the original, clean state $y_0$.

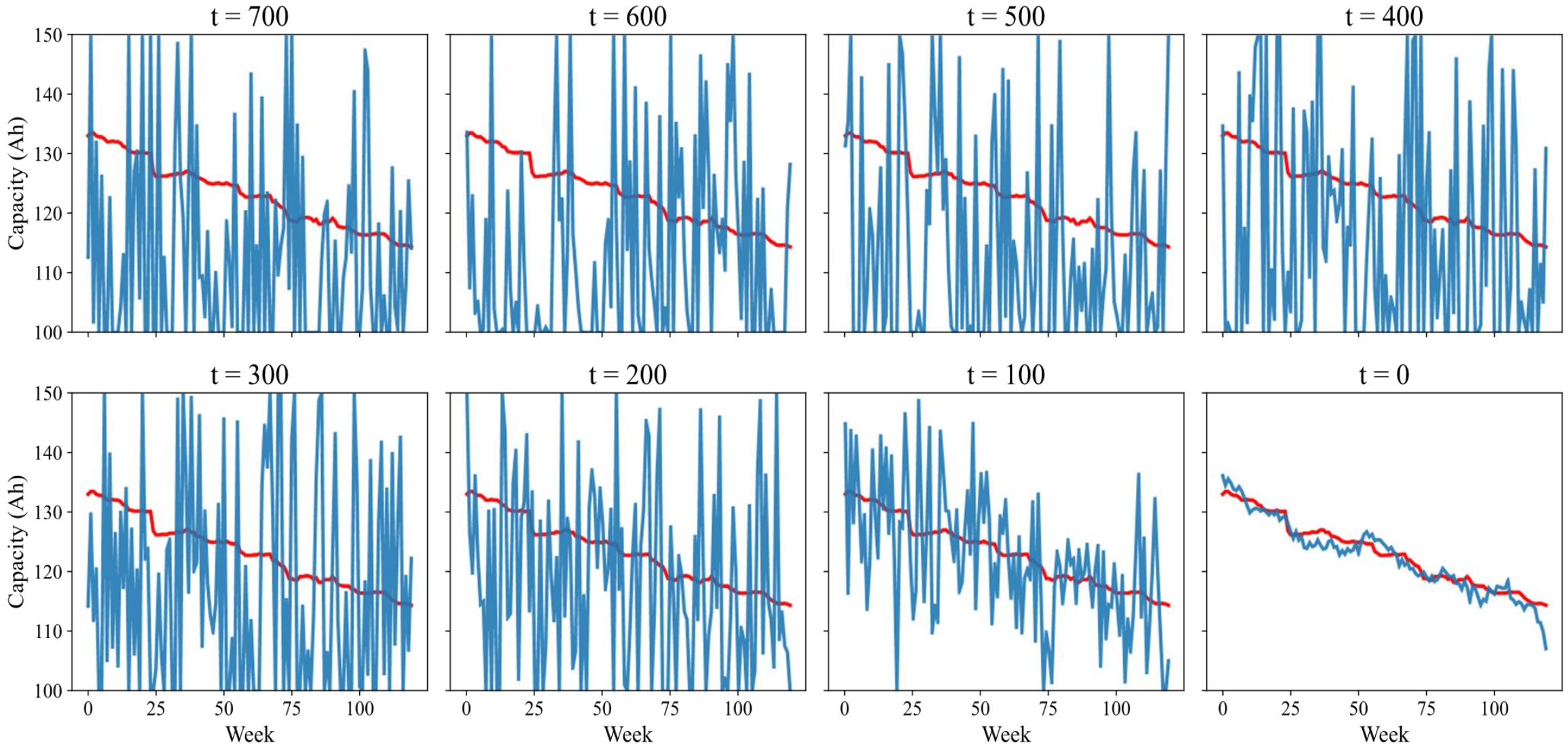

Fig. 4. Visualization of the reverse denoising process.

This figure illustrates how the model starts from a random noise sample at timestep t=700 and progressively removes the predicted noise to generate a clean battery capacity degradation trajectory at t=0. In the CDUA model, this denoising process is conditional. It is guided by the features of the historical battery data $x$, and its probability distribution is approximated by a neural network model:

$$p_\theta(y_{t-1} \mid y_t, x) = \mathcal{N}(\mu_\theta(y_t, t, x), \Sigma_\theta(y_t, t, x)) \tag{5}$$

Theoretically, we could directly learn both the mean and the covariance. However, according to the research on DDPM, when $\beta_t$ is small, the covariance can be approximated as a fixed value $\beta_t I$. This simplifies the task, as the model only needs to predict the mean, which in turn can be represented by a neural network's prediction of the noise. The formula for the mean is defined as:

$$\mu_\theta(y_t, t, x) = \frac{1}{\sqrt{\alpha_t}}\left( y_t - \frac{\beta_t}{\sqrt{1-\alpha_t}} \epsilon_\theta(y_t, t, x) \right) \tag{6}$$

Thus, the core task of the model is simplified to training a neural network to predict the noise added to $y_t$ . This network, incorporating the historical data features x as a condition, learns to reverse the diffusion process and generate realistic capacity sequences.

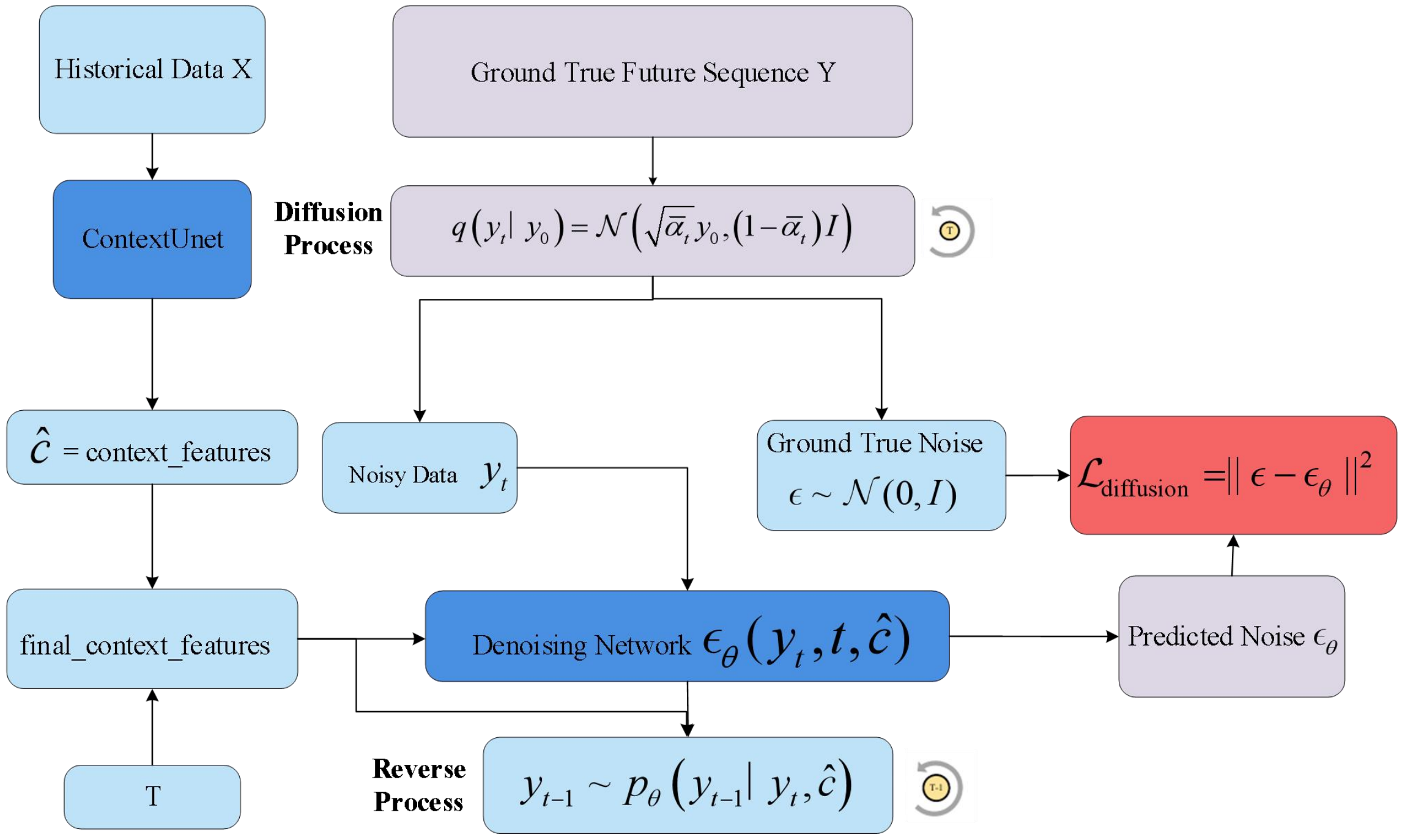

Fig. 5. CDUA model training and inference process.

### 3.2 CDUA model architecture

The CDUA model's core principle is to use the model's ability to generate data distributions to create future capacity sequences by progressively denoising data, all while being guided by features from historical data. This process, illustrated in Fig. 5, is split into two main parts. First, a context feature extractor, built on a U-Net architecture, processes historical battery data to create a feature map rich in contextual information. This feature map serves as a crucial condition for the denoising process. Second, the central component of the diffusion model is the noise predictor; it receives three inputs: the noisy future sequence, the embedded representation of the timestep, and the contextual feature map from the extractor. By learning to predict and remove the noise added to the future sequence, the predictor effectively reverses the diffusion process. The model is trained by minimizing the error between the predicted noise and the actual noise, ensuring it accurately learns to recover the true future sequence distribution from noise. Finally, during the inference stage, the model starts with a pure noise sequence and uses the trained noise predictor to iteratively reverse the diffusion process, gradually generating a future capacity sequence that aligns with the predicted distribution, thereby effectively quantifying the prediction's uncertainty.

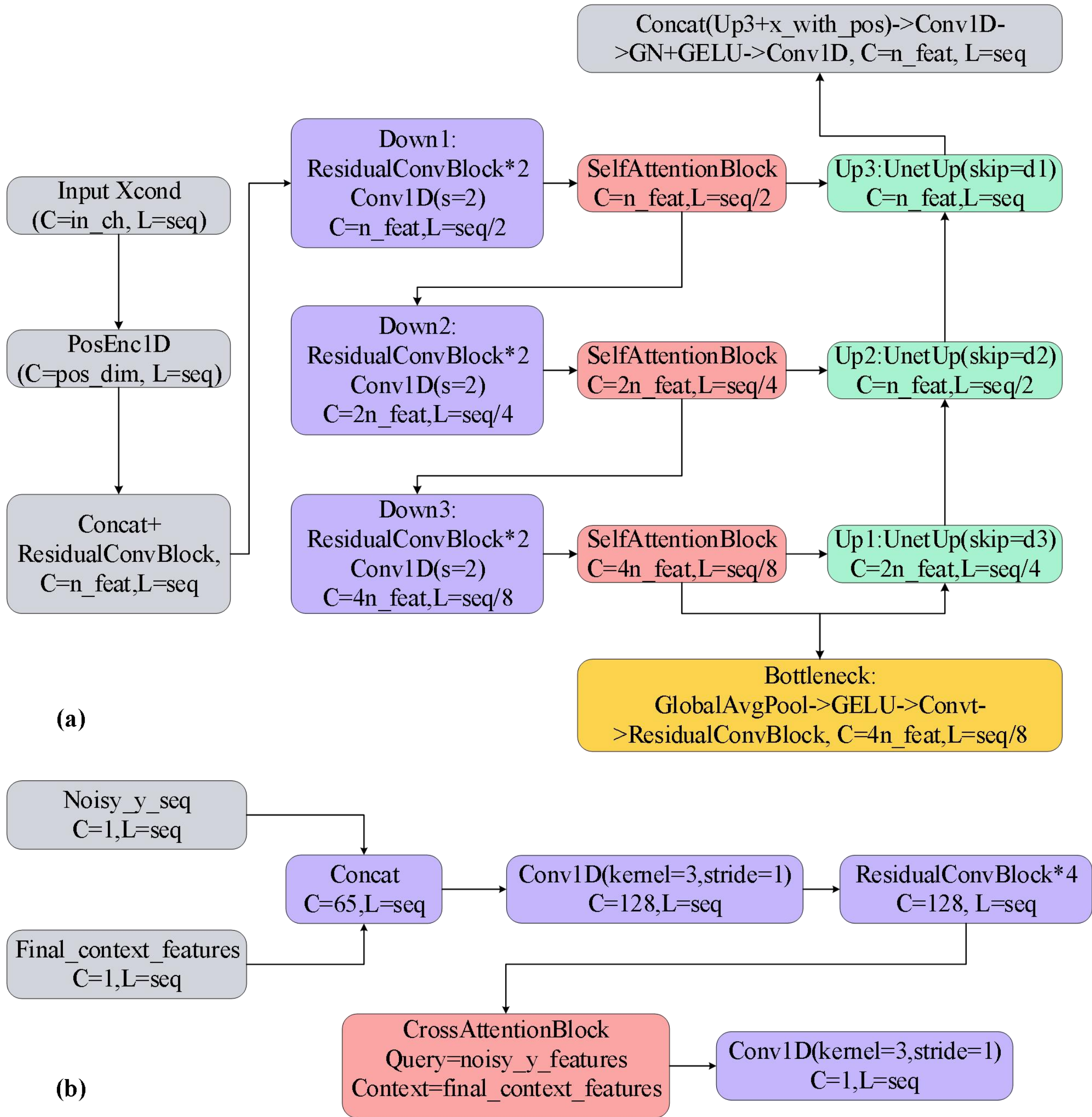

Fig. 6. Detailed architecture of the CDUA model components. (a) architecture of the ContextUnet in the CDUA model, (b) architecture of the noise predictor in the CDUA model.

### 3.3 Module design

The context feature extractor uses a U-Net architecture to process one-dimensional time series data, mirroring its original design for medical image segmentation. Fig. 6(a) shows the details of ContextUnet. The network's structure includes a down-sampling path that progressively shrinks the input data. This is done through a series of stages, each featuring two Residual Convolutional Blocks to prevent vanishing gradients, followed by a 1D convolutional layer that halves the sequence length. At the end of each down-sampling stage, a Self-Attention Block is added. This lets the model focus on global interactions within the data, creating a more comprehensive contextual representation than a purely local approach could. This path captures high-level features before the data reaches the bottleneck layer, which consolidates the most abstract information from the encoder. Furthermore, a sinusoidal positional encoding is added to the input features to inject information about the relative or absolute position of the data points in the sequence, which is crucial for the model to understand temporal order.

From the bottleneck, the model transitions to the up-sampling path to restore the original sequence length. This process involves multiple stages where data is up-sampled using transposed convolutions or linear interpolation. The key to this phase is the use of skip connections. These connections link the high-level features from the encoder directly to the decoder's low-level features. This fusion ensures that the model can combine the broad, semantic information from the deeper layers with the fine-grained details preserved in the earlier layers. This prevents the loss of critical timing information during down-sampling and results in a more precise reconstruction of the output sequence.

The noise predictor is a central component of the diffusion model, designed to predict and remove noise during the reverse process. It is shown in the Fig. 6(b). A cross-attention mechanism is introduced within this module to facilitate the effective fusion of conditional information from historical data with the noisy sequence. Specifically, the noisy sequence serves as the query, while the contextual features from the U-Net extractor act as the key and value. This design allows the noise predictor to selectively focus on the most relevant historical information to guide its predictions. By leveraging this mechanism, the model ensures that the generated output remains consistent with the provided contextual information, producing predictions that are both accurate and aligned with the underlying data dynamics. Furthermore, the timestep embedding vector is also integrated into this process by being added to the context feature map, enabling the noise predictor to make more informed decisions by considering both the historical context and the current denoising stage.

### 3.4 Training and inference

The primary objective of training the CDUA model is to enable the neural network to predict the noise as accurately as possible, making it closely match the actual noise added during the forward diffusion process. This work uses the Mean Squared Error (MSE) as the loss function to quantify this difference:

$$L_{\mathrm{MSE}} = \| \quad - \quad {}_{\theta}(y_t, t, \hat{c}) \|^2 \tag{7}$$

During training, we apply a masking technique to the sequences, ensuring that the loss is calculated only on the valid data points and not on the padded, meaningless parts of the sequence. This approach prevents invalid data from interfering with the model's learning process. The model's parameters are updated through gradient descent to minimize this loss, effectively teaching the model how to reverse the diffusion process and remove noise conditioned on the historical data. To ensure the reproducibility of our results, the model is trained and evaluated using a specific set of hyperparameters. The key parameters for the diffusion process, model architecture, and training optimization are detailed in Table 1.

Table 1. Key hyperparameters for the CDUA model.

| Hyperparameter | Value |
| --- | --- |
| Diffusion Timesteps (T) | 700 |

| Beta Schedule | Linear |
|---|---|
| Epochs | 1000 |
| Batch Size | 16 |
| Optimizer | Adam |
| Cross-Validation Folds | 4 |
| Sampling Trajectories (N) | 40 |
| Confidence Interval (CI) | 95% |

After the model is trained, we perform probabilistic forecasting by running the full reverse denoising process. The sampling procedure begins with a completely random sequence of Gaussian noise. From there, we iterate backward from t=700 down to t=0. In each step, the trained model predicts the noise, which is then used to update the sequence and progressively remove the noise. The update rule for generating a cleaner sequence $y_{t-1}$, from the noisy sequence $y_t$, is defined as:

$$y_{t-1} = \frac{1}{\sqrt{\bar{\alpha}_t}} \left( y_t - \frac{1-\bar{\alpha}_t}{\sqrt{1-\alpha_t}} \epsilon_\theta(y_t, t, \hat{c}) \right) + \sigma_t Z \tag{8}$$

where $Z$ is a random sample from a standard normal distribution, and $\sigma_t$ is a small, predefined noise value. Because a random noise vector is introduced in each sampling step, repeating this process multiple times will yield a unique future capacity trajectory for each run. To capture the inherent uncertainty, we repeat the sampling process N times (in this study, N = 40) using the same historical data. These N trajectories collectively form a predictive distribution of the future capacity. From this distribution, we can derive both point forecasts and quantify uncertainty, the final point forecast for each time step is obtained by taking the average of the N generated trajectories at that specific time point. The model's uncertainty is quantified by calculating the standard deviation of the N trajectories at each time step. This standard deviation is then used to construct confidence intervals. For a 95% confidence interval, the upper and lower bounds are calculated as $\text{mean} \pm 1.96 \times \text{std}$ . This method provides a clear and interpretable range for the model's predictions, reflecting the inherent variability and randomness of the underlying process.

**4. Results and discussion**

To evaluate the performance of the proposed CDUA model, this chapter details a series of experiments and presents an in-depth, multi-faceted analysis. We first introduce the key metrics used for performance evaluation and outline the overall structure of this section. Subsequently, we conduct a detailed discussion from three core perspectives. First, we compare the performance of the CDUA model against two mainstream baseline models (LSTM and Seq2Seq) to assess its advantages in both point prediction accuracy and uncertainty quantification. Second, a feature ablation study is conducted to validate the effectiveness of our proposed hybrid feature engineering framework. Third, a model ablation study is performed to dissect the contributions of the core components within the CDUA model, namely the self-attention and cross-attention mechanisms. Finally, this chapter summarizes all experimental findings to highlight the comprehensive superiority of our model.

For a robust assessment, we adopted three key metrics. We use RMSE (Root Mean Square Error) and MAE (Mean Absolute Error) to measure the accuracy of point predictions, and the 95% confidence interval width to evaluate the precision of uncertainty quantification. The RMSE is more sensitive to large errors due to its squared penalty term, thereby better reflecting the model's robustness. It is defined as:

$$\text{RMSE} = \sqrt{\frac{1}{N}\sum_{i=1}^{N}(y_i - \hat{y}_i)^2} \tag{9}$$

The MAE, with its linear penalty, provides a more intuitive measure of the average error magnitude. Its formula is:

$$\text{MAE} = \frac{1}{N}\sum_{i=1}^{N}|y_i - \hat{y}_i| \tag{10}$$

Meanwhile, the confidence interval width quantifies the range of prediction uncertainty, provided that the prediction interval coverage is maintained. A smaller confidence interval width value indicates a more precise prediction of uncertainty. It is calculated as:

$$\text{CI Width} = \frac{1}{N}\sum_{i=1}^{N}(U_i - L_i) \tag{11}$$

where N is the number of samples, $U_i$ and $L_i$ represent the upper and lower bounds of the predicted confidence interval for the $i_{th}$ sample, respectively. Together, these metrics enable a thorough and deep analysis of the model's overall performance from the dual perspectives of point prediction accuracy and interval prediction precision.

#### 4.1 Performance comparison with baseline models

The proposed CDUA model is compared with two widely-used baseline models for time series forecasting, the LSTM network and the Seq2Seq model. All models are evaluated on a dataset comprising data from 20 real-world electric vehicles, employing a rolling-forecast strategy. Specifically, historical sequences of 8, 16, 24, and 32 time steps are used to predict the subsequent battery capacity sequence.

In terms of point prediction accuracy, the CDUA model demonstrates a significant advantage, as shown in Table II. This superiority is particularly pronounced in short-term forecasting scenarios. For instance, at an 8-week horizon, the CDUA model achieves a relative RMSE of 1.14% and a relative MAE of 0.94%. These figures are substantially lower than those of both LSTM (relative RMSE of 1.33%, relative MAE of 1.09%) and Seq2Seq (relative RMSE of 1.24%, relative MAE of 1.03%). Specifically, our model reduces the relative RMSE by approximately 14.3% compared to LSTM and 8.1% compared to Seq2Seq. This indicates that the diffusion-based architecture is more adept at capturing short-term dependencies within the data, thereby yielding more precise short-range forecasts.

Table 2. Performance comparison of point prediction accuracy.

| Model | RMSE | MAE |
|---|---|---|

| | 8 | 16 | 24 | 32 | 8 | 16 | 24 | 32 |
|---|---|---|---|---|---|---|---|---|
| LSTM | 1.33% | 1.35% | 1.38% | 1.38% | 1.09% | 1.11% | 1.15% | 1.16% |
| Seq2Seq | 1.24% | 1.29% | 1.35% | 1.35% | 1.03% | 1.06% | 1.12% | 1.11% |
| CDUA | 1.14% | 1.31% | 1.22% | 1.33% | 0.94% | 1.10% | 1.01% | 1.12% |

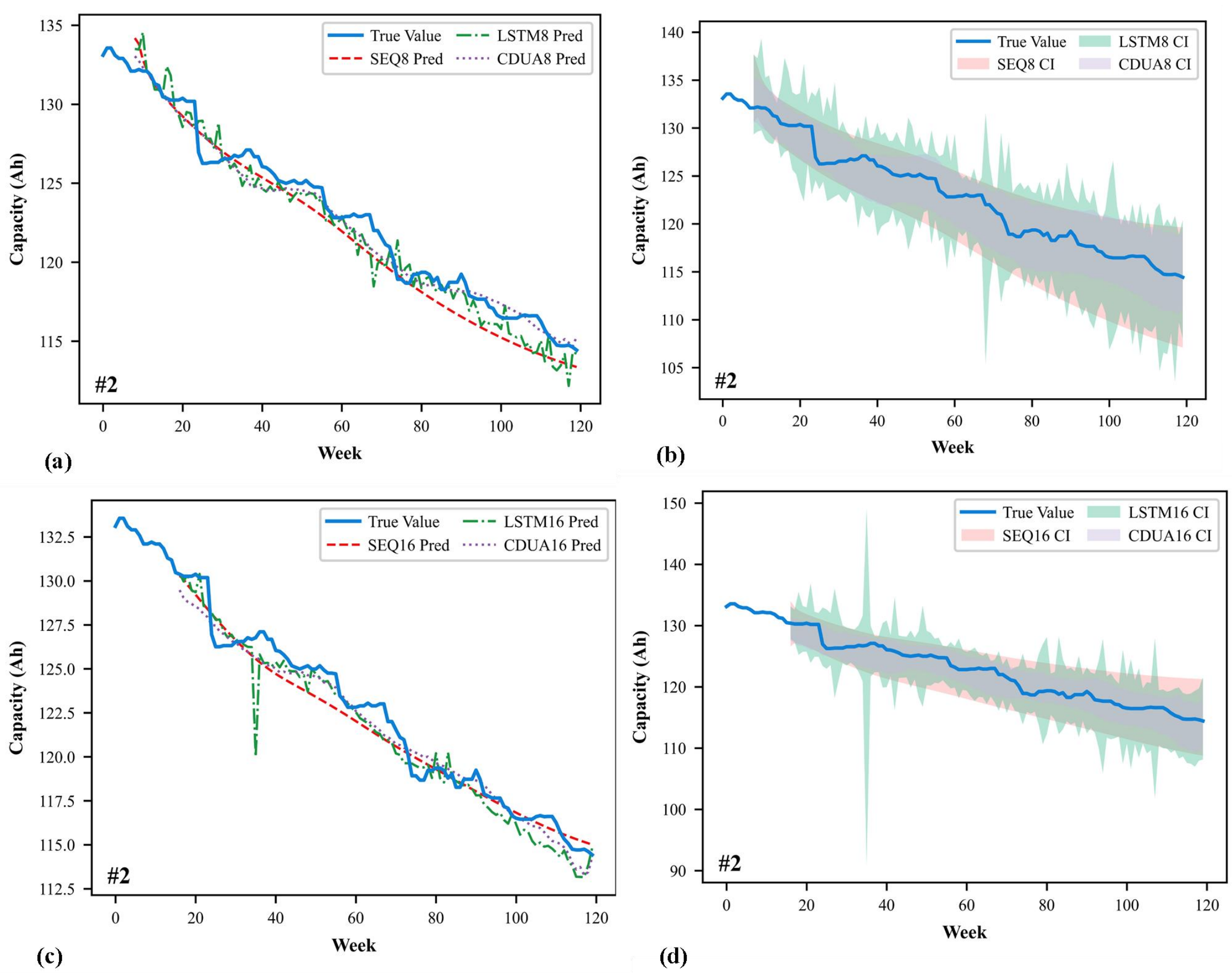

Fig. 7. Visualization of prediction performance comparison between the CDUA model and baselines on vehicle #2. (a) battery capacity point prediction based on 8-Week input, (b) battery capacity interval prediction based on 8-Week input, (c) battery capacity point prediction based on 16-Week input. (d) battery capacity interval prediction based on 16-Week input.

The CDUA model also excels in uncertainty quantification. Crucially, it is important to note that all competing models successfully maintained a Prediction Interval Coverage Probability (PICP) of approximately 93%, which is close to the nominal 95% target, thereby ensuring a fair and valid comparison of the interval widths. With this reliability established, the superiority of our model becomes evident in its significantly narrower 95% Confidence Interval Width, as detailed in Table III. For instance, at the 8-week forecast horizon, the CDUA model's relative CI Width is only 3.74%. This is substantially narrower than both LSTM's (6.40%) and Seq2Seq's (5.36%), representing a reduction in relative confidence interval width of approximately 41.6% and 30.2%, respectively. This powerfully demonstrates that the diffusion-based architecture can generate highly precise uncertainty estimates without sacrificing coverage, providing a more reliable foundation for risk assessment and decision-making.

To provide a more intuitive demonstration of the model's predictive capabilities, Fig. 7 visualizes a comparison of the long-term prediction results from the CDUA model and the baselines on vehicle #2. The plots clearly show that the CDUA model's prediction trajectory aligns most accurately with the ground truth capacity degradation curve across different historical input lengths. It successfully captures the complex non-linearities and local fluctuations inherent in the real-world data, whereas the LSTM and Seq2Seq models exhibit overly smoothed trajectories, leading to larger cumulative errors over the prediction horizon. Furthermore, the 95% confidence intervals, visualized as shaded areas, highlight the model's exceptional capability in uncertainty quantification. The CDUA model consistently produces remarkably narrow confidence interval width that tightly bound the true values, signifying well-calibrated and confident predictions. In stark contrast, the confidence interval widths of the baseline models are considerably wider, indicating substantial predictive uncertainty and thus diminishing their reliability and practical utility in mission-critical applications. This visual evidence compellingly substantiates the dual superiority of our CDUA model in both the accuracy of its point predictions and the precision of its uncertainty quantification.

Table 3. Comparison of 95% confidence interval width.

| Model | CI Width(8) | CI Width(16) | CI Width(24) | CI Width(32) |
|---|---|---|---|---|
| LSTM | 6.40% | 6.67% | 6.66% | 6.72% |
| Seq2Seq | 5.36% | 5.75% | 5.48% | 6.23% |
| CDUA | **3.74%** | **3.61%** | **3.67%** | **3.82%** |

### 4.2 Feature ablation study

To investigate the impact of different input features on the model's performance, we conduct a comprehensive feature ablation study. As detailed in Section II, our feature engineering process involved a hybrid strategy. Specifically, we designed three distinct feature sets: F1 (Baseline Features) comprises 6 core features selected based on Pearson correlation analysis; F2 (Auxiliary Features) contains 4 supplementary features identified using XGBoost's feature importance ranking; and F3 (Comprehensive Features) is the final feature set used in our proposed model, formed by the union of F1 and F2, totaling 9 unique features. We train and evaluate our model separately using each of these feature sets, with the performance metrics summarized in Table IV. The results unequivocally demonstrate that the comprehensive feature set F3 achieves the best overall performance. For instance, in short-term prediction (8-week horizon), F3 obtains the lowest relative RMSE of 1.14% and relative MAE of 0.94%. Furthermore, it consistently yields a narrower relative CI Width across most horizons, with the best result of 3.61% at the 16-week horizon, indicating superior uncertainty quantification.

Fig. 8 provides a detailed visual analysis of the model's performance, showcasing the results for a representative test fold from our 5-fold cross-validation. To ensure a robust evaluation across the 20-sample dataset, each test set consists of four unique samples, which are differentiated by the four marker styles within each plot.

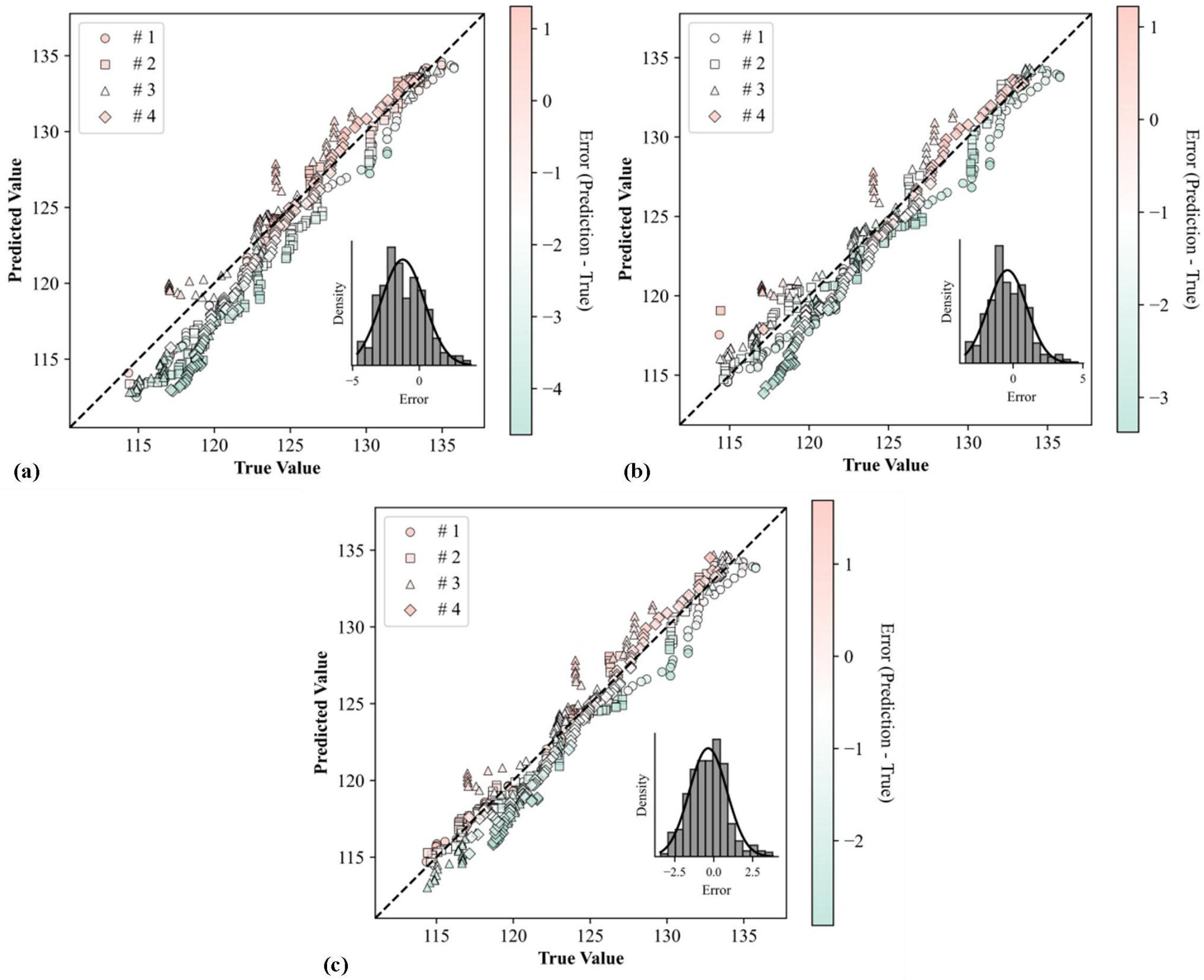

Fig. 8. Scatter plots of predicted vs. true values for feature ablation study. The results are shown for feature sets (a) F1, (b) F2, and (c) F3. The inset histograms show the distribution of prediction errors.

The comparison across feature sets on these test samples is stark. The plot for feature set F3 shows data points that are significantly more clustered around the diagonal line. Concurrently, its error distribution histogram is narrower and more sharply centered at zero. This provides strong visual evidence that the F3 feature set enables the model to learn a more generalizable relationship, resulting in more accurate and reliable predictions across a diverse set of test samples.

Table 4. Results of the feature ablation study.

| Feature set | Metric | 8 | 16 | 24 | 32 |
| --- | --- | --- | --- | --- | --- |
| | RMSE | 1.36% | 1.38% | 1.24% | 1.34% |

| F1(6 features) | MAE | 1.12% | 1.15% | 1.05% | 1.12% |
|---|---|---|---|---|---|
| | CI Width | 3.81% | 4.04% | 3.99% | 4.08% |
| | RMSE | 1.24% | 1.29% | 1.33% | 1.32% |
| F2(4 features) | MAE | 1.03% | 1.10% | 1.11% | 1.10% |
| | CI Width | 3.66% | 3.79% | 3.95% | 3.84% |
| | RMSE | 1.14% | 1.31% | 1.22% | 1.33% |
| F3(9 features) | MAE | 0.94% | 1.10% | 1.01% | 1.12% |
| | CI Width | 3.74% | 3.61% | 3.67% | 3.82% |

This outcome shows that the F1 and F2 feature sets are complementary, not redundant. Each set appears to capture different data patterns. F1 captures more fundamental trends, while F2 captures more complex ones. Combining them into F3 gives the model a more complete dataset, which improves its ability to learn temporal patterns and estimate uncertainty. This confirms the success of our hybrid approach to feature selection.

### 4.3 Model ablation study

To validate the contributions of our model's core components, we conduct a model ablation study. We compare the performance of our full CDUA model against three ablated variants: a Backbone model lacking both attention mechanisms, CDUA with/without(w/o) Self-Attention, and CDUA w/o Cross-Attention. The results, presented in Table V, unequivocally demonstrate that both attention mechanisms are crucial for the model's performance, as the full CDUA model consistently outperforms all ablated versions. The removal of the Self-Attention mechanism (CDUA w/o Self-Attention) proved particularly detrimental, causing the most significant performance decay; for instance, the relative RMSE in 8-week ahead prediction surged from 1.14% to 1.30%. This underscores its critical role in capturing long-range temporal dependencies. Similarly, removing the Cross-Attention mechanism (CDUA w/o Cross-Attention) also results in a substantial drop in performance (relative RMSE increasing to 1.22%), highlighting its indispensable function in integrating information across different feature dimensions. Unsurprisingly, the Backbone model yielded the worst results, confirming the foundational importance of these modules.

The box plots in Fig. 9 intuitively illustrate the absolute prediction error distributions of the different model variants. Compared to the ablated models, the box plot for the full CDUA model consistently features a shorter interquartile range and a lower median line across all prediction horizons. This visually signifies that the complete model's prediction error distribution is more concentrated around a lower median, indicating a reduced probability of large errors. Collectively, these findings prove that the Self-Attention and Cross-Attention mechanisms are not merely additive but work synergistically as integral parts of the CDUA architecture. In particular, the significant performance degradation upon removing Self-Attention underscores its vital role in capturing complex temporal patterns from the limited real-world data (20 vehicles), validating our approach for small-sample scenarios.

Table 5. Performance comparison from the model ablation study.

| Model Variant | Metric | 8 | 16 | 24 | 32 |
|---|---|---|---|---|---|
| Backbone | RMSE | 1.40% | 1.37% | 1.24% | 1.36% |
| | MAE | 1.09% | 1.16% | 1.04% | 1.13% |
| | CIWidth | 3.82% | 3.62% | 3.76% | 3.51% |
| CDUA w/o Self-Attention | RMSE | 1.30% | 1.33% | 1.28% | 1.33% |
| | MAE | 1.09% | 1.12% | 1.05% | 1.11% |
| | CIWidth | 3.73% | 3.71% | 3.65% | 3.60% |
| CDUA w/o Cross-Attention | RMSE | 1.22% | 1.29% | 1.22% | 1.32% |
| | MAE | 1.03% | 1.08% | 1.03% | 1.11% |
| | CIWidth | 3.74% | 3.45% | 3.83% | 3.62% |
| CDUA | RMSE | 1.14% | 1.31% | 1.22% | 1.33% |
| | MAE | 0.94% | 1.10% | 1.01% | 1.12% |
| | CIWidth | 3.74% | 3.61% | 3.67% | 3.82% |

In summary, this section has validated the superiority of our proposed CDUA model through a holistic and multi-faceted experimental evaluation. We have not only demonstrated its superior accuracy in point prediction and precision in uncertainty quantification against established baselines but have also, through rigorous ablation studies, confirmed the synergistic and indispensable roles of our chosen features and attention mechanisms. These findings collectively establish a robust experimental and theoretical foundation for deploying uncertainty-aware, high-precision predictive models in complex real-world systems.

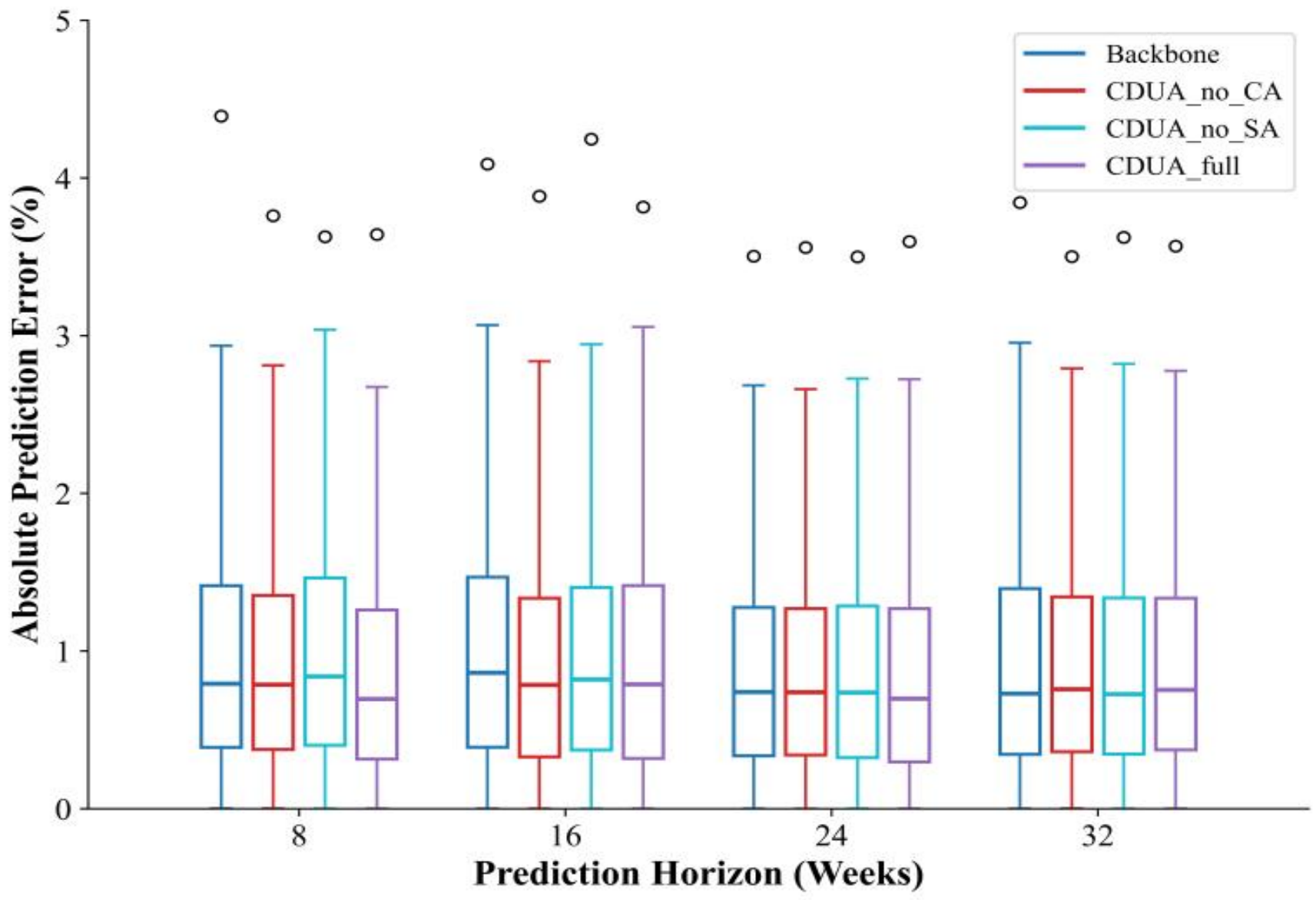

Fig. 9. Box plots of absolute prediction errors for the model ablation study.

## 5. Conclusion

To address the challenge of high-precision capacity prediction for lithium-ion batteries, this paper proposes the CDUA model. This method integrates the powerful generative capabilities of diffusion models with the precise capture of temporal dependencies afforded by attention mechanisms, effectively resolving the dilemma of insufficient accuracy and reliability faced by traditional methods under complex operating conditions.

This study first extracts highly correlated key features from real-time vehicle data through systematic feature engineering, ensuring the validity of the model's input. Subsequently, these features are utilized to drive the CDUA model, which consists of a contextual U-Net and a noise predictor network. The contextual U-Net leverages a self-attention mechanism to deeply encode the complex temporal patterns of the input features, providing rich contextual guidance for the noise predictor network. Under this guidance, the noise predictor network estimates the noise present in the signal, which is then used to accurately recover the target capacity value step-by-step.

Experimental results validate the superior performance of the proposed method in terms of both high accuracy and reliability. On a real-world vehicle operation dataset, the CDUA model achieves a relative MAE and RMSE as low as 0.94% and 1.14%, respectively, demonstrating significant superiority over other mainstream methods. Furthermore, another key advantage of this method lies in its effective quantification of prediction uncertainty. The model achieves a narrow 95% confidence interval width of 3.74% while maintaining a high coverage rate. This signifies that it can not only provide precise point estimates but also clearly delineate the reliability bounds of its predictions. This capability is of paramount importance for safety-critical Battery Management Systems and provides a more solid foundation for subsequent intelligent decision-making.

## Acknowledgement

This work is supported in part by the National Natural Science Foundation of China (Grant No. 52472401) and Natural Science Foundation of Xinjiang Uygur Autonomous Region (Grant No. 2024D01C14). The authors gratefully acknowledge the great help of the fund.

## Data and code availability

The code will be made publicly available on GitHub after the publication.